\documentclass{article}

\usepackage[final]{neurips_2021}

\usepackage[utf8]{inputenc} 
\usepackage[T1]{fontenc}    
\usepackage{hyperref}       
\usepackage{url}            
\usepackage{booktabs}       
\usepackage{amsfonts}       
\usepackage{amssymb}
\usepackage{amsmath}
\usepackage{nicefrac}       
\usepackage{microtype}      
\usepackage{xcolor}         
\usepackage[ruled,vlined]{algorithm2e}
\usepackage{subcaption}
\usepackage{wrapfig}
\usepackage{caption}
\usepackage{graphicx}
\usepackage{float}
\usepackage{enumitem}
\usepackage[export]{adjustbox}
\setlist[enumerate]{itemsep=0mm}

\SetKwInput{KwInput}{Input}
\DeclareMathOperator*{\argmin}{argmin}
\DeclareMathOperator*{\argmax}{argmax}

\newtheorem{definition}{Definition}

\title{Model based Multi-agent Reinforcement Learning with Tensor Decompositions}

\author{Pascal Van Der Vaart \\ TU Delft \\ \texttt{p.r.vandervaart-1@tudelft.nl} \And Anuj Mahajan \\ University of Oxford \\ \texttt{anuj.mahajan@cs.ox.ac.uk} \And Shimon Whiteson \\ University of Oxford \\ \texttt{shimon.whiteson@cs.ox.ac.uk}}

\begin{document}

\maketitle

\begin{abstract}
A challenge in multi-agent reinforcement learning is to be able to generalize over intractable state-action spaces. Inspired from Tesseract~\citep{mahajan2021tesseract}, this position paper investigates generalisation in state-action space over unexplored state-action pairs by modelling the transition and reward functions as tensors of low CP-rank. Initial experiments on synthetic MDPs show that using tensor decompositions in a model-based reinforcement learning algorithm can lead to much faster convergence if the true transition and reward functions are indeed of low rank.
\vspace{-3mm}
\end{abstract}

\section{Introduction}
\vspace{-3mm}
Recent progress in multi-agent RL has been promising towards creating agents which are capable of generalising over multiple tasks \citep{oel2021}, they have also demonstrated effectiveness in dealing with the problem of exploration in a large action space \citep{mahajan2019maven, gupta2020uneven} and overcoming intractability arising from its exponential growth in the number of agents\citep{mahajan2021tesseract, wang2020rode, wang2020qplex} when learning under constraints like decentralisation. Inspired from Tesseract~\citep{mahajan2021tesseract}, which utilises tensor decomposition structure in factored action spaces, we investigate whether tensor decompositions can be used to attain generalisation across the state-action space in cooperative multi-agent setting towards obtaining better sample efficiency. In this position paper, we focus on the model based setting. Initial empirical results on randomly generated MDPs provide promising evidence for state-action generalisation and sample efficiency using tensor decompositions over baseline model based algorithms which do not use the tensor approximation.

In multi-agent reinforcement learning, the goal is to find a policy for multiple agents that performs well on a given task. Tasks are formalized by Markov decision processes (MDP), which are described by a transition function and a reward function. In the RL setting, the transition and reward functions are unknown, and finding a policy that achieves high reward requires exploration to gather data from the MDP and learn its dynamics. Because the size of the action space grows exponentially with the amount of agents, it is especially important in multi-agent reinforcement learning to learn with high sample efficiency as coverage over all state action pairs is not feasible. This work showcases how the use of low rank CP-decompositions can drastically improve sample efficiency of classic model-based reinforcement learning algorithms and provide better generalisation when the transition and reward tensors are of low CP-rank.

\section{Background}
\vspace{-3mm}
\subsection{Tensor decompositions}
\vspace{-2mm}
\begin{definition}\label{definition:tensor}
An order $n$ tensor over a field $\mathbb{F}$ with dimensions $d_1, d_2, \dots, d_n$ is a multilinear map
$
    T: \mathbb{F}^{d_1} \times \mathbb{F}^{d_2} \times \cdots \times \mathbb{F}^{d_n} \rightarrow \mathbb{F},
$
and can be represented by a n-dimensional $d_1 \times d_2 \times \cdots \times d_n$ array $T_{i_1i_2\dots i_n}$ such that the mapping is defined as
\[
T(u^1, \dots, u^n) = \sum_{i_1=1}^{d_1}\sum_{i_2=1}^{d_2}\dots\sum_{i_n=1}^{d_n} T_{i_1i_2\dots i_n} u^1_{i_1} u^2_{i_2} \dots u^n_{i_n}
\]
for $u^i \in \mathbb{F}^{d_i}$.
For a set of indices $i_1, \dots i_k \in \{1, \dots, n\}$ the expression $$T(u^1, \dots, u^{i_1-1}, I, u^{i_1+1}, \dots u^{i_2 -1}, I, u^{i_2 + 1}, \dots, u^{i_k - 1}, I, u^{i_k +1}, \dots, u^n)$$ denotes a $d_{i_1} \times \dots \times d_{i_k}$ tensor defined by the mapping $(u^{i_1}, u^{i_2}, \dots, u^{i_k}) \mapsto T(u^1, \dots, u^n)$. 

The set of order n tensors with dimensions $d_1, d_2, \dots d_n$ over $\mathbb{F}$ is denoted by $\mathbb{F}^{d_1 \times d_2 \times \cdots \times d_n}$.
\end{definition}

The CANDECOMP/PARAFAC (CP) decomposition for tensors can be thought of as a generalization of the singular value decomposition for matrices. Related work involving tensors and tensor decompositions can be found in appendix \ref{section:relatedwork}.
\begin{definition}\label{definition:CPdecomp}
A rank $r$ CP-decomposition of a tensor $T \in \mathbb{F}^{d_1 \times d_2 \times \cdots \times d_n}$ is a set of vectors $ (u_l^i)^{i = 1, \dots n}_{l=1, \dots r}, u_i^l \in \mathbb{F}^{d_i}$ and scalars $(w_l)_{l=1, \dots, r} \in \mathbb{F}$ such that 

\[
    T = \sum_{l=1}^r w_l u_l^1 \otimes u_l^2 \otimes \cdots \otimes u_l^n
,\]
where $\|u_l^i\| = 1$ for all $l \in \{1, \dots, r\}$ and $i \in \{1, \dots, n\}$.

The tensor $T$ is said to be of CP-rank $r$ if $r$ is the smallest number for which a rank $r$ CP-decomposition for $T$ exists.
\end{definition}

It is clear that if a large $n \times n \times n$ tensor $T$ has low rank, the search space of an application which requires an estimate of $T$ can be greatly reduced by incorporating the low rank information. Instead of estimating $n^3$ parameters, the problem can be described by $3rn$ parameters in decomposed form instead. The main idea of this work is to use this fact to efficiently estimate the transition and reward tensors in discrete multi-agent reinforcement learning problems.

\subsection{Reinforcement learning}
\vspace{-2mm}
In reinforcement learning, the goal is to compute a strategy to perform a certain task. Tasks are formalized as Markov decision processes (MDPs) $(S, A, T, R, \gamma)$, where $S$ and $A$ are the state and action spaces, $T$ and $R$ are the transition and reward function and $\gamma$ is the discount factor. At each time step $t \in \mathbb{N}$ an agent chooses an action $a_t \in A$ based on the state $s_t \in S$. The environment then returns a reward $r_t = R(s_t, a_t)$ and the next state $s_{t+1} \sim T( \cdot | s_t, a_t)$. The strategy to choose actions is called the policy $\pi: S \times A \rightarrow [0, 1]$ which defines a probability distribution over the actions given the current state.
The goal of the agent is to maximize the expected discounted reward $\mathbb{E}_{T, \pi}\left[\sum_{t=1}^H \gamma^t r_t\right]$, where the expectation is over the states and actions, whose probability distributions are implied by the transition function $T$ and policy $\pi$.

In multi-agent reinforcement learning (MARL), there are multiple agents that interact with the environment as opposed to only one agent. Each agent has its own action space $A_i$, which means that the transition function is now a function $S \times A_1 \times \cdots \times A_n \times S \rightarrow [0, 1]$ and the reward function is $S \times A_1 \times A_2 \times \dots \times A_n \rightarrow [0, 1]$.


Clearly this can be cast as a single agent reinforcement learning problem by setting $A = A_1 \times \cdots \times A_n$. A result of this is that the action space grows exponentially large with the number of agents, further increasing the requirement of efficient exploration. Instead of casting it as a single agent reinforcement learning problem, explicitly incorporating the multi-agent paradigm allows to exploit more structure in the MDP.

In this work, this is done by considering the transition function to be a tensor $T \in \mathbb{R}^{S \times A_1 \times \cdots \times A_n \times S}$ such that $T_{sa_1\dots a_n s'} = T(s, a_1, \dots, a_n, s')$. The reward function is analogously written as a tensor $R \in \mathbb{R}^{s\times a_1 \times \cdots \times a_n}$. If the tensors $T$ and $R$ are of low rank, models formed by an agent during training can be expected to generalize across unseen state-action pairs.
\section{Methods}
\vspace{-3mm}
\subsection{Tensor decomposition algorithms}
\vspace{-2mm}
While computing tensor decompositions is NP-hard in general \cite{hillarnphard}, there exist algorithms such as \cite{harshman70} and \cite{animaaltmin} which are proven to converge in special cases. The algorithm used in this work is an ablation of the alternating rank 1 updates algorithm presented in \cite{animaaltmin}. The restarts, clustering and clipping procedures are left out to form a shorter and simpler algorithm which still performs well in practice. The algorithm as used is presented in algorithms  \ref{algo:deflation}, \ref{algo:altmin}, and \ref{algo:altrank1updates}. The main idea of the algorithm is to run asymmetric power updates to compute a good starting value for alternating minimization, which further improves the accuracy of the decomposition.

\begin{algorithm}[b]
\SetAlgoLined
\KwInput{A tensor $T \in \mathbb{R}^{m \times n \times p}$ and decomposition rank $r$}
\KwResult{$\{u_k^j\}_{k=1\dots r}^{j=1 \dots n}, \{w_k\}_{k=1\dots r}$ such that $T \approx \sum_{k=1}^r w_k u_k^1 \otimes u_k^2 \otimes \cdots \otimes
u_k^n$}
    $(\{u^j_k\}_{k=1\dots r}^{j=1 \dots n}, \{w_k\}_{k=1\dots r} )= \text{PowerIteration}(T, r)$ (algorithm \ref{algo:deflation})\;
    $(\{u^j_k\}_{k=1\dots r}^{j=1 \dots n}, \{w_k\}_{k=1\dots r})= \text{AlternatingMinimize}(T, \{u^j_k\}_{k=1\dots r}^{j=1 \dots n}, \{w_k\}_{k=1\dots r})$ (algorithm \ref{algo:altmin})\;
 \caption{Alternating rank 1 updates}\label{algo:altrank1updates}
\end{algorithm}

\subsection{Tensor completion}
\vspace{-2mm}
In the tensor completion problem, the goal is to recover a tensor with only partially observed entries. 
For a tensor $T \in \mathbb{R}^{d_1 \times \cdots \times d_n}$, let $\Omega \in \{0, 1\}^{d_1 \times \cdots \times d_n}$ denote a mask such that $\Omega_{i_1\dots i_n} = 1$ if and only if entry $T_{i_1 \dots i_n}$ has been observed. A method proposed in \cite{tensorcompletion1} and \cite{tensorcompletion2} involves solving the minimization problem 
 $$\argmin_{\{u_k^j\}_{k=1\dots r}^{j=1 \dots n} \in \mathbb{R}^{d_j}, \{w_k\}_{k=1\dots r} \in \mathbb{R}} \|\Omega \cdot T - \Omega \cdot \sum_{k=1}^r w_k u_k^1 \otimes u_k^2 \otimes \cdots \otimes u_k^n\|_F,$$
where $\cdot$ denotes an entrywise multiplication. Algorithm \ref{algo:altrank1updates} can be used to solve this problem with a slight modification to the alternating minimization step as showcased in appendix \ref{section:appendixtensorcompletion}.

\subsection{Model based reinforcement learning}
\vspace{-2mm}
In model based reinforcement learning agents make models of the environment to plan ahead, instead of attempting to maximize reward directly. Under the assumption that the transition and reward tensors are of low rank, using tensor decomposition allows for sample efficient models that generalize over unseen state-action pairs.

The deterministic reward tensor is estimated using tensor completion, where the unobserved entries are simply the state-action pairs the agents have never experienced. After enough exploration, enough entries of the reward tensor will be revealed to reconstruct the entire tensor.

The algorithm presented in this paper follows a very standard model-based reinforcement learning approach and is presented in algorithm \ref{algo:modelbasedrldecomp}. The \texttt{NORMALIZE} function is an entry wise division so that the resulting tensor is a transition tensor, that is the sum over the resulting states is 1. The \texttt{POLICYIMPROVEMENT} function is clarified in appendix \ref{section:policyimprovement}. 


\subsection{Relationship to Tesseract}\label{section:relationsshiptesseract}
\vspace{-2mm}
This method differs from \cite{mahajan2021tesseract} because in this work, decompositions of the entire tensor $T \in \mathbb{R}^{S \times A_1 \times \cdots \times A_n \times S}$ and $R \in \mathbb{R}^{S \times A_1 \times \cdots \times A_n}$ are computed. Model based Tesseract instead considers for each $s, s' \in S$ the $A_1 \times \cdots \times A_2$ tensor $\tilde T_{ss'} = T(e_s, I, \dots, I, e_s')$, and computes an individual tensor decomposition for each state and next state pair. Analogously, it considers for each state $s \in S$ the reward tensor $R_s = R(e_s, I, \dots, I)$ and computes a decomposition for every state.

In theory, both methods can represent the same transition and reward functions. To see this, consider for example an MDP with 2 agents, such that the reward tensor is of order 3. Let $R = \sum_{i=1}^r w_i x_i \otimes y_i \otimes z_i$ be the true reward tensor. This can represented by Tesseract by setting $R_s = \sum_{i=1}^r w_i \langle e_s, x_i \rangle y_i \otimes z_i$, where $e_s$ is the $s$-th standard basis vector. This results in the combined reward tensor $$\sum_{s \in S} e_s \otimes R_s = \sum_{s \in S} e_s \otimes \sum_{i=1}^r w_i \langle e_s, x_i \rangle y_i \otimes z_i =  \sum_{i=1}^r w_i \sum_{s \in S} e_s \otimes \langle e_s, x_i \rangle y_i \otimes z_i = R$$

Conversely, if $R$ is of the form $\sum_{s \in S} e_s \otimes R_s$ where each $R_s$ is of rank $r$, then the rank of $R$ is bounded by $|S|r$ so it can be represented in our framework. Thus, when low rank structure spans across states, out method would ensure better sample efficiency as it would require fewer number of parameters.
\begin{algorithm}\caption{CP-Decomposed state-action space reinforcement learning}
    \KwInput{An MDP, state space size $S$, action space sizes $A_1, \dots A_n$, Hyperparameters: $n_{\texttt{episodes}}, n_{\texttt{train}}, \epsilon, n_\texttt{improvement iter}$}
    
    \KwResult{Policy $\pi$ with good performance}
    
    $D = 0 \in \mathbb{N}^{S \times A_1 \times \cdots \times A_n \times S}$\;
    $R = 0 \in \mathbb{R}^{S \times A_1 \times \cdots \times A_n}$\;
    Initialize random $\pi$\;
    
    \For{$\texttt{episode} = 1, 2, \dots, n_\texttt{episodes}$}{
        $s_1 \sim MDP$\;
        $a_1 \sim \begin{cases}
            \pi(s_0) &\text{ with probability } 1 - \epsilon(\texttt{episode}) \\
            \texttt{Unif}(A) &\text{ with probability } \epsilon(\texttt{episode})
        \end{cases}$\;
        \For{$t = 1, \dots, \texttt{episode length}$}{
            $(s_{t+1}, r_t) \sim MDP( . | s_t, a_t)$\;
            $D_{s_ta_ts_{t+1}} = D_{s_ta_ts_{t+1}} + 1$\;
            $R_{s_ta_t} = r_t$\;
            
            $a_t \sim \begin{cases}
                \pi(s_t) &\text{ with probability } 1 - \epsilon(\texttt{episode})\\
                \texttt{Unif}(A) &\text{ with probability } \epsilon(\texttt{episode})
            \end{cases}$\;
        }
        \If{$n_\texttt{episodes}\,\%\,n_\texttt{train} = 0$}{
            $\hat T = \texttt{DECOMP}(\texttt{NORMALIZE}(D), r_T)$\;
            $\Omega = D > 0$\;
            $\hat R = \texttt{TENSORCOMPLETION}(R, \Omega, r_R)$\;
            $\pi = \texttt{POLICYIMPROVEMENT}(\pi, \hat T, \hat R, n_\texttt{improvement iter})$\;
        }
    }
\end{algorithm}\label{algo:modelbasedrldecomp}
\vspace{-3mm}
\section{Experiments}
\vspace{-3mm}
\subsection{Random transition and reward functions of predefined rank}
\vspace{-2mm}
This experiment involves algorithm \ref{algo:modelbasedrldecomp} applied to an MDP described by a randomly generated transition tensor $T$ and reward tensor $R$. The MDP has 20 states, and 3 agents with 10 actions each, leading to 20000 state-action pairs. Both the transition and reward tensors are of rank $5$. More information on how they are generated can be found in appendix \ref{section:tensorgeneration}. 


We tested three different agents for experiments. The first agent is a baseline agent which uses no decompositions. It uses the maximum likelihood estimator for $T$ and fills in missing rewards for unvisited state-action pairs with the mean of the visited rewards.
Secondly, an agent using decomposition across entire state action space for for $T$ and $R$ with three settings of approximate rank $5$ (exact), $3$ (insufficient), $10$ (overparametrised). The final agent is model based Tesseract with rank $5$ and $1$ decompositions. Note that the rank $5$ case can represent the correct transition and reward tensors, but is overparametrised for the task (60000 parameters versus 350 for our rank $5$ agent for the transition function). Similarly, the rank $1$ agent will be insufficient for representing the actual dynamics but will provide faster learning. 

For each agent, if the slice through the estimated transition tensor $T$ corresponding to a specific state-action pair contains only zeros, all entries are set to $\frac{1}{|S|}$. This means that if there is no estimate for $T( \cdot | s, a_1, a_2, a_3)$, a uniform distribution is assumed instead. This happens for the no decompositions agent exactly when a state-action pair has never been visited before.

\begin{figure}
    \centering
    \includegraphics[width=\textwidth]{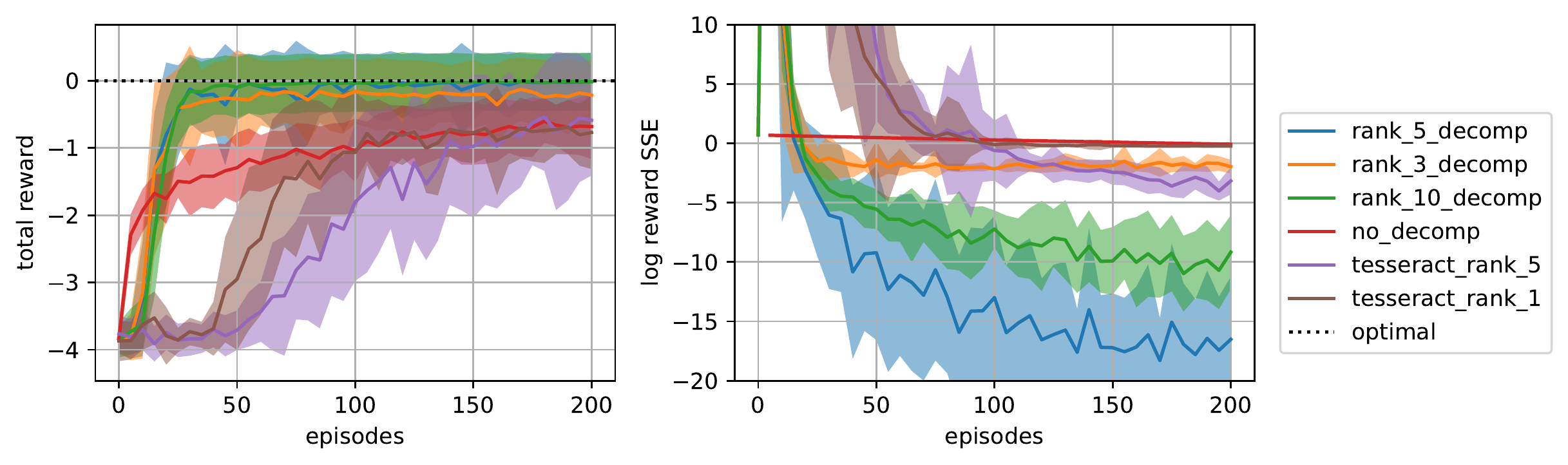}
    \caption{Total reward per episode at test time (left) and sum of squared errors of the reward tensor estimate (right) after a number of episodes of training.}
    \label{fig:predef_rewards&reward}
\end{figure}

\begin{figure}
    \centering
    \includegraphics[width=\textwidth]{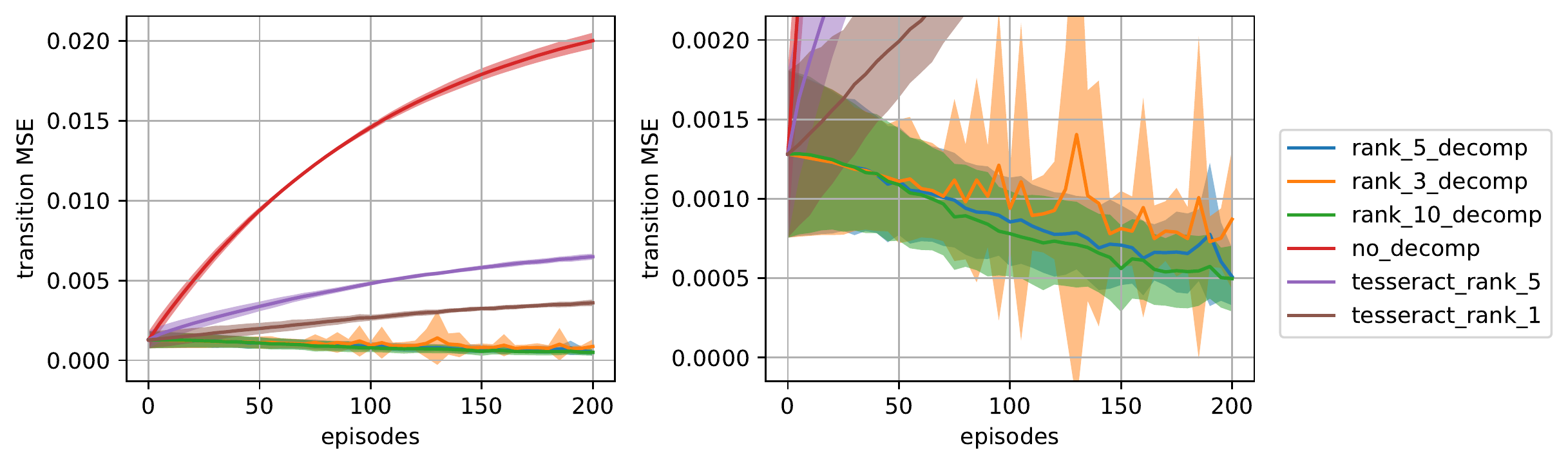}
    \caption{Mean squared error of the estimate of the transition tensor after training a number of episodes in the first experiment. The right plot is a zoomed in version of the left plot.}
    \label{fig:predef_transitionerrors}
    \vspace{-6mm}
\end{figure}

Each agent is trained for 200 episodes, recomputing their models and applying policy improvement every 10 episodes. The agents use $\epsilon$ greedy exploration with epsilon decaying from $0.9$ to $0.1$. During training, the total episodic rewards, errors in the transition tensor and errors in the reward tensors are tracked.
The entire experiment is ran 20 times, with newly generated $T$ and $R$ for each run. The optimal reward in each experiment is computed beforehand via policy improvement on the true functions $T$ and $R$, and then for each experiment the optimal reward is subtracted from the episodic rewards so that optimal performance is a reward of $0$ for each experiment. Finally, the number of unique visited state-action pairs is also tracked. The results of the experiment are shown in figures \ref{fig:predef_rewards&reward}, \ref{fig:predef_transitionerrors} and \ref{fig:predef_visited}. Figure \ref{fig:predef_rewards&reward} shows that algorithm \ref{algo:modelbasedrldecomp} significantly outperforms a standard model-based approach without tensor decompositions in the setting where $T$ and $R$ are of low rank. While the agent with rank 3 decompositions achieves a slightly sub-optimal policy, the performance seems to be quite robust against incorrectly guessing the correct rank for the problem. 

Interestingly, agents without tensor decompositions outperform the agents that use tensor decompositions during the first few episodes. This can be attributed to the unrobustness of tensor completion. Figure \ref{fig:predef_rewards&reward} shows that for our algorithm, during the first 20 episodes the error in the reward tensor can be of order $10^5$ and higher, because the optimization problem is very ill-conditioned when little entries are revealed. Tesseract suffers even more from this problem, as each individual state now requires sufficiently many revealed entries. A way to overcome these problems could be for example to take the naive estimate without tensor completion when attempting tensor completion results in very extreme values, or adding regularization to the optimization problem. Figure \ref{fig:predef_rewards&reward} also shows that with sufficiently many revealed entries, our method achieves very good approximates of the reward tensor. If the approximate reward tensor rank is set correctly (rank 5), the reward tensor is recovered almost exactly after visiting only 4000 (see figure \ref{fig:predef_visited}) or 20\% of the state-action pairs. Setting the rank results in slower convergence, but still yields a reasonably good estimate. Finally, setting the rank too low causes the agent to be incapable of representing the true reward tensor, but on limited revealed entries this estimate still outperforms the estimate without decompositions. Tesseract with rank 5 decompositions takes a long time to get a good estimate, but eventually outperforms our method with rank 3 decompositions. This is explained by the analysis in  \ref{section:relationsshiptesseract}, which showed that rank 5 Tesseract is in theory capable to represent the true reward tensor, albeit requiring many more samples in comparison as confirmed by this experiment.

Figure \ref{fig:predef_transitionerrors} shows the error in transition tensors. Note that the error of Tesseract and the agent without decompositions increases over time. This is due to the fact that for many states-action pairs, the default uniform distribution assigning probability $\frac{1}{|S|}$ to each state is a better estimate than an extreme distribution resulting from only one observation of that state-action pair. Figure \ref{fig:predef_visited} shows that even after 200 episodes, only around 11000 state action pairs out of 20000 total are visited, meaning that many state-action pairs are likely to have been visited only once. This means that unless an agent can combine information from different state-action pairs, it is unfeasible to make a good transition function estimate. Since using no decompositions assumes every state-action pair to be independent, there is no generalization across states-action pairs. Tesseract does slightly better as it attempts to generalize the action space for each state independently, but figure \ref{fig:predef_transitionerrors} shows that our method produces significantly better transition tensor estimates by attempting to generalize over the combined state-action space.
\vspace{-3mm}

\subsection{MDP with degenerate states}
\begin{figure}
    \vspace{-3mm}
    \centering
    \includegraphics[width=\textwidth]{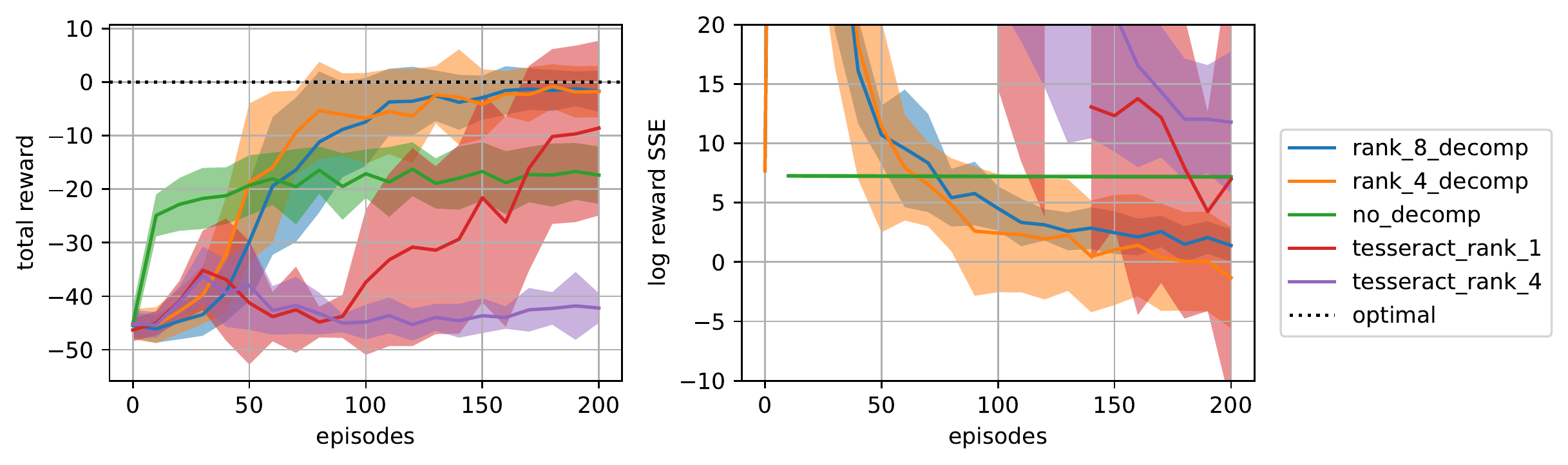}
    \caption{Total reward per episode at test time (left) and sum of squared errors of the reward tensor estimate (right) after a number of episodes of training in the state degeneracy MDP.}
    \label{fig:degen_rewards&reward}
    \vspace{-3mm}
\end{figure}

\vspace{-2mm}
In this experiment we test algorithm \ref{algo:modelbasedrldecomp} on state degeneracy, a situation where our method can provide further sample efficiency. State degeneracy can occur when observations are noisy. We use an MDP with 3 agents, this time with 16 states and each agent has an action space of size 20. The 16 states are split into 4 groups, where each group has the same transition function of rank 1, and a linearly dependent reward tensor of rank 1. This means that rank 1 Tesseract is expected to be able to recover exact models after enough iterations. Furthermore, similar to the analysis in \ref{section:relationsshiptesseract}, writing $R = \sum_{i=1, 5, 9, 13} (e_i + e_{i+1} + e_{i+2} + e_{i+3}) \otimes R_i$ where $R_i$ denotes a reward tensor for each group, reveals that the entire reward tensor is of rank at most 4. Similar to the previous experiment, we consider the following agents: A baseline agent using no decomposition, Agents using decomposition across state action space with ranks $4$ and $8$, Tesseract with rank $4$ and $1$.
The entire experiment is repeated 20 times. The results are shown in figures \ref{fig:degen_rewards&reward}, \ref{fig:degen_transitionerrors} and \ref{fig:degen_visited}. Like in the results of the first experiment, figure \ref{fig:degen_rewards&reward} shows that the agents that use decompositions accross state-action space outperform the other agents in terms of total reward obtained. This is mostly attributable to the performance on the reward tensor error. We also observe that Tesseract is unable to recover a good reward tensor estimate in the given sample budget. In contrast to the previous experiment, the transition tensor estimates of the agents that use state-action decompositions do not differ significantly from the estimates made by Tesseract. This can be explained by the fact that in this experiment, the groups themselves have entirely independent transition and reward functions, which means that generalization is only possible within groups. In contrast, the low rank structure imposed on the entire transition tensor in the first experiment allowed our method to generalize over all states.
\vspace{-3mm}

\section{Conclusion}
\vspace{-3mm}
In this position paper we investigated whether tensor decompositions can be used across state actions space for better sample efficiency in RL for the model based setting. Our experiments show that an algorithm which computes CP-decompositions of the environment models has significant advantages when the MDP is described by low rank transition and reward functions.
\newpage
\bibliography{bibliography}

\begin{thebibliography}{22}
\providecommand{\natexlab}[1]{#1}
\providecommand{\url}[1]{\texttt{#1}}
\expandafter\ifx\csname urlstyle\endcsname\relax
  \providecommand{\doi}[1]{doi: #1}\else
  \providecommand{\doi}{doi: \begingroup \urlstyle{rm}\Url}\fi

\bibitem[Mahajan et~al.(2021)Mahajan, Samvelyan, Mao, Makoviychuk, Garg,
  Kossaifi, Whiteson, Zhu, and Anandkumar]{mahajan2021tesseract}
Anuj Mahajan, Mikayel Samvelyan, Lei Mao, Viktor Makoviychuk, Animesh Garg,
  Jean Kossaifi, Shimon Whiteson, Yuke Zhu, and Animashree Anandkumar.
\newblock Tesseract: Tensorised actors for multi-agent reinforcement learning.
\newblock In \emph{Proceedings of the 38th International Conference on Machine
  Learning}, volume 139, pages 7301--7312. PMLR, 2021.
\newblock URL \url{https://proceedings.mlr.press/v139/mahajan21a.html}.

\bibitem[DeepMind-OEL et~al.(2021)DeepMind-OEL, Stooke, Mahajan, Barros, Deck,
  Bauer, Sygnowski, Trebacz, Jaderberg, Mathieu, McAleese, Bradley-Schmieg,
  Wong, Porcel, Raileanu, Hughes-Fitt, Dalibard, and Czarnecki]{oel2021}
DeepMind-OEL, Adam Stooke, Anuj Mahajan, Catarina Barros, Charlie Deck, Jakob
  Bauer, Jakub Sygnowski, Maja Trebacz, Max Jaderberg, Michael Mathieu, Nat
  McAleese, Nathalie Bradley-Schmieg, Nathaniel Wong, Nicolas Porcel, Roberta
  Raileanu, Steph Hughes-Fitt, Valentin Dalibard, and Wojciech~Marian
  Czarnecki.
\newblock Open-ended learning leads to generally capable agents.
\newblock \emph{arXiv preprint arXiv:2107.12808}, 2021.

\bibitem[Mahajan et~al.(2019)Mahajan, Rashid, Samvelyan, and
  Whiteson]{mahajan2019maven}
Anuj Mahajan, Tabish Rashid, Mikayel Samvelyan, and Shimon Whiteson.
\newblock Maven: Multi-agent variational exploration.
\newblock In \emph{Advances in Neural Information Processing Systems}, pages
  7611--7622, 2019.

\bibitem[Gupta et~al.(2020)Gupta, Mahajan, Peng, B{\"o}hmer, and
  Whiteson]{gupta2020uneven}
Tarun Gupta, Anuj Mahajan, Bei Peng, Wendelin B{\"o}hmer, and Shimon Whiteson.
\newblock Uneven: Universal value exploration for multi-agent reinforcement
  learning.
\newblock \emph{arXiv preprint arXiv:2010.02974}, 2020.

\bibitem[Wang et~al.(2020{\natexlab{a}})Wang, Gupta, Mahajan, Peng, Whiteson,
  and Zhang]{wang2020rode}
Tonghan Wang, Tarun Gupta, Anuj Mahajan, Bei Peng, Shimon Whiteson, and
  Chongjie Zhang.
\newblock Rode: Learning roles to decompose multi-agent tasks.
\newblock \emph{arXiv preprint arXiv:2010.01523}, 2020{\natexlab{a}}.

\bibitem[Wang et~al.(2020{\natexlab{b}})Wang, Ren, Liu, Yu, and
  Zhang]{wang2020qplex}
Jianhao Wang, Zhizhou Ren, Terry Liu, Yang Yu, and Chongjie Zhang.
\newblock Qplex: Duplex dueling multi-agent q-learning.
\newblock \emph{arXiv preprint arXiv:2008.01062}, 2020{\natexlab{b}}.

\bibitem[Hillar and Lim(2013)]{hillarnphard}
Christopher~J. Hillar and Lek-Heng Lim.
\newblock Most tensor problems are np-hard.
\newblock \emph{J. ACM}, 60\penalty0 (6), November 2013.
\newblock ISSN 0004-5411.
\newblock \doi{10.1145/2512329}.
\newblock URL \url{https://doi.org/10.1145/2512329}.

\bibitem[Harshman(1970)]{harshman70}
Richard Harshman.
\newblock Foundations of the parafac procedure: Models and conditions for an
  "explanatory" multi-modal factor analysis.
\newblock \emph{UCLA Working Papers in Phonetics}, 16, 1970.

\bibitem[Anandkumar et~al.(2015)Anandkumar, Ge, and Janzamin]{animaaltmin}
Animashree Anandkumar, Rong Ge, and Majid Janzamin.
\newblock Guaranteed non-orthogonal tensor decomposition via alternating
  rank-$1$ updates, 2015.

\bibitem[Jain and Oh(2014)]{tensorcompletion1}
Prateek Jain and Sewoong Oh.
\newblock Provable tensor factorization with missing data.
\newblock In Z.~Ghahramani, M.~Welling, C.~Cortes, N.~Lawrence, and K.~Q.
  Weinberger, editors, \emph{Advances in Neural Information Processing
  Systems}, volume~27. Curran Associates, Inc., 2014.
\newblock URL
  \url{https://proceedings.neurips.cc/paper/2014/file/c15da1f2b5e5ed6e6837a3802f0d1593-Paper.pdf}.

\bibitem[Liu and Moitra(2020)]{tensorcompletion2}
Allen Liu and Ankur Moitra.
\newblock Tensor completion made practical.
\newblock In H.~Larochelle, M.~Ranzato, R.~Hadsell, M.~F. Balcan, and H.~Lin,
  editors, \emph{Advances in Neural Information Processing Systems}, volume~33,
  pages 18905--18916. Curran Associates, Inc., 2020.
\newblock URL
  \url{https://proceedings.neurips.cc/paper/2020/file/dab1263d1e6a88c9ba5e7e294def5e8b-Paper.pdf}.

\bibitem[Anandkumar et~al.(2012)Anandkumar, Hsu, and
  Kakade]{anandkumar2012method}
Animashree Anandkumar, Daniel Hsu, and Sham~M. Kakade.
\newblock A method of moments for mixture models and hidden markov models.
\newblock In Shie Mannor, Nathan Srebro, and Robert~C. Williamson, editors,
  \emph{Proceedings of the 25th Annual Conference on Learning Theory},
  volume~23 of \emph{Proceedings of Machine Learning Research}, pages
  33.1--33.34, Edinburgh, Scotland, 25--27 Jun 2012. PMLR.
\newblock URL \url{https://proceedings.mlr.press/v23/anandkumar12.html}.

\bibitem[Anandkumar et~al.(2014)Anandkumar, Ge, Hsu, Kakade, and
  Telgarsky]{anandkumar2014latent}
Animashree Anandkumar, Rong Ge, Daniel Hsu, Sham~M. Kakade, and Matus
  Telgarsky.
\newblock Tensor decompositions for learning latent variable models.
\newblock \emph{J. Mach. Learn. Res.}, 15\penalty0 (1):\penalty0 2773–2832,
  January 2014.
\newblock ISSN 1532-4435.

\bibitem[Cichocki et~al.(2017)Cichocki, Phan, Zhao, Lee, Oseledets, Sugiyama,
  and Mandic]{Cichocki2017TensorNF}
Andrzej Cichocki, A.~Phan, Qibin Zhao, Namgil Lee, I.~Oseledets, Masashi
  Sugiyama, and Danilo~P. Mandic.
\newblock Tensor networks for dimensionality reduction and large-scale
  optimization: Part 2 applications and future perspectives.
\newblock \emph{Found. Trends Mach. Learn.}, 9:\penalty0 431--673, 2017.

\bibitem[Cheng et~al.(2017)Cheng, Wang, Zhou, and Zhang]{cheng2017}
Yu~Cheng, Duo Wang, Pan Zhou, and Tao Zhang.
\newblock A survey of model compression and acceleration for deep neural
  networks.
\newblock \emph{CoRR}, abs/1710.09282, 2017.
\newblock URL \url{http://arxiv.org/abs/1710.09282}.

\bibitem[Kossaifi et~al.(2019)Kossaifi, Bulat, Tzimiropoulos, and
  Pantic]{Kossaifi2019TNetPF}
Jean Kossaifi, Adrian Bulat, Georgios Tzimiropoulos, and Maja Pantic.
\newblock T-net: Parametrizing fully convolutional nets with a single
  high-order tensor.
\newblock \emph{2019 IEEE/CVF Conference on Computer Vision and Pattern
  Recognition (CVPR)}, pages 7814--7823, 2019.

\bibitem[Kossaifi et~al.(2020)Kossaifi, Toisoul, Bulat, Panagakis, Hospedales,
  and Pantic]{Kossaifi2020FactorizedHC}
Jean Kossaifi, Antoine Toisoul, Adrian Bulat, Yannis Panagakis, Timothy~M.
  Hospedales, and Maja Pantic.
\newblock Factorized higher-order cnns with an application to spatio-temporal
  emotion estimation.
\newblock \emph{2020 IEEE/CVF Conference on Computer Vision and Pattern
  Recognition (CVPR)}, pages 6059--6068, 2020.

\bibitem[Bulat et~al.(2020)Bulat, Kossaifi, Tzimiropoulos, and
  Pantic]{Bulat2020IncrementalML}
Adrian Bulat, Jean Kossaifi, Georgios Tzimiropoulos, and Maja Pantic.
\newblock Incremental multi-domain learning with network latent tensor
  factorization.
\newblock In \emph{AAAI}, 2020.

\bibitem[Sunehag et~al.(2018)Sunehag, Lever, Gruslys, Czarnecki, Zambaldi,
  Jaderberg, Lanctot, Sonnerat, Leibo, Tuyls, and Graepel]{sunehag2018}
Peter Sunehag, Guy Lever, Audrunas Gruslys, Wojciech~Marian Czarnecki, Vinicius
  Zambaldi, Max Jaderberg, Marc Lanctot, Nicolas Sonnerat, Joel~Z. Leibo, Karl
  Tuyls, and Thore Graepel.
\newblock Value-decomposition networks for cooperative multi-agent learning
  based on team reward.
\newblock In \emph{Proceedings of the 17th International Conference on
  Autonomous Agents and MultiAgent Systems}, AAMAS '18, page 2085–2087,
  Richland, SC, 2018. International Foundation for Autonomous Agents and
  Multiagent Systems.

\bibitem[Rashid et~al.(2018)Rashid, Samvelyan, Schroeder, Farquhar, Foerster,
  and Whiteson]{rashid2018qmix}
Tabish Rashid, Mikayel Samvelyan, Christian Schroeder, Gregory Farquhar, Jakob
  Foerster, and Shimon Whiteson.
\newblock {QMIX}: Monotonic value function factorisation for deep multi-agent
  reinforcement learning.
\newblock In Jennifer Dy and Andreas Krause, editors, \emph{Proceedings of the
  35th International Conference on Machine Learning}, volume~80 of
  \emph{Proceedings of Machine Learning Research}, pages 4295--4304. PMLR,
  10--15 Jul 2018.
\newblock URL \url{https://proceedings.mlr.press/v80/rashid18a.html}.

\bibitem[Bromuri(2012)]{tensorizedqlearning}
Stefano Bromuri.
\newblock A tensor factorization approach to generalization in multi-agent
  reinforcement learning.
\newblock In \emph{2012 IEEE/WIC/ACM International Conferences on Web
  Intelligence and Intelligent Agent Technology}, volume~2, pages 274--281,
  2012.
\newblock \doi{10.1109/WI-IAT.2012.21}.

\bibitem[Azizzadenesheli et~al.(2016)Azizzadenesheli, Lazaric, and
  Anandkumar]{azizzadenesheli2016}
Kamyar Azizzadenesheli, Alessandro Lazaric, and Animashree Anandkumar.
\newblock Reinforcement learning of pomdps using spectral methods.
\newblock 06 2016.

\end{thebibliography}
\appendix
\section{Related work}\label{section:relatedwork}
Work on tensor decompositions in general machine learning include \cite{anandkumar2012method}, which uses the CP-decomposition to learn mixture models and hidden Markov models, and \cite{anandkumar2014latent} learns latent variable models. Another application of tensor methods is to compress neural networks in \cite{Cichocki2017TensorNF} and \cite{cheng2017}. \cite{Kossaifi2019TNetPF} parametrizes convolutional nets with a high-order tensor of low rank to reduce over-parameterization, with applications to spatio-temporal tasks in \cite{Kossaifi2020FactorizedHC}. In \cite{Bulat2020IncrementalML} this parametrization is used for multi-domain image classification.

Previous reinforcement learning methods that attempt to exploit structure in the multi-agent setting include VDN \cite{sunehag2018}, which models the joint Q-function (see appendix \ref{section:policyimprovement}) as a sum of the agents individual Q-functions. This is generalised by QMIX \cite{rashid2018qmix}, which learns a monotonic function of the individual Q-functions instead of taking a sum.

Methods for generalization in multi agent reinforcement learning using specifically tensor decompositions include \cite{tensorizedqlearning}, where tensor decompositions are used to factorize the $Q$-function in model free $Q-$learning algorithms. Our method instead factorizes the estimated transition and reward functions of the MDP in a model based algorithm. This idea was initially proposed in \cite{mahajan2021tesseract}, which contains a model free algorithm and a model based algorithm. The difference between the model based algorithm in \cite{mahajan2021tesseract} and our method, is that we factorize over the state-action space, while \cite{mahajan2021tesseract} factorizes only over the action space. This allows our work to potentially generalize over unseen states instead of only over unseen actions. 

Work in tensor decompositions for partially observable MDPs (POMPDs) in a single agent setting include \cite{azizzadenesheli2016}. Adapting our method for generalisation in multi agent MDPs for POMPDs is an interesting future research direction

\section{Decomposition algorithms}
\begin{algorithm}
\SetAlgoLined
\KwInput{A tensor $T \in \mathbb{R}^{d_1 \times \cdots \times d_n}$ and decomposition rank $r$, tolerance $\epsilon$}
\KwResult{Vectors $\{u_k^j\}_{k=1\dots r}^{j=1 \dots n}$ and scalars $\{w_k\}_{k=1\dots r}$ such that $T \approx \sum_{k=1}^r w_k u_k^1 \otimes u_k^2 \otimes \cdots \otimes u_k^n$}

 \For{k = 1, \dots r}{
    \For{j = 1, \dots, n}{
        $u_{k, 0}^j \sim \mathcal{N}(0, 1)$\;
        $u_{k, 0}^j = \frac{u_k^j}{\|u_k^j\|}$\;
    }
    \While{$ \sum_{j=1}^n \|u_{k, n+1}^j - u^j_{k, n}\|^2 > \epsilon$}{
        \For{j=1, \dots, n}{
            $u^{j}_{k, m+1} = \frac{T(u^1_{k, m}, \dots, u^{j-1}_{k, m}, I, u^{j+1}_{k, m}, \dots, u^n_{k, m} )}{\|T(u^1_{k, m}, \dots, u^{j-1}_{k, m}, I, u^{j+1}_{k, m}, \dots, u^n_{k, m} )\|}$\;
        }
    }
    $w_k = T(u_k^1, \dots, u_k^n)$\;
    $w_k^N = c_kw_k^N$\;
    $T = T - w_k u_k^1 \otimes u_k^2 \otimes \cdots \otimes u_k^n$\;
 }
\caption{Tensor power iteration with deflation}\label{algo:deflation}
\end{algorithm}

\begin{algorithm}[H]
\SetAlgoLined
\KwInput{A tensor $T$, starting values for $\{u_k^j\}_{k=1\dots r}^{j=1 \dots n}, \{w_k\}_{k=1\dots r}$}
\KwResult{$\{u_k^j\}_{k=1\dots r}^{j=1 \dots n}, \{w_k\}_{k=1\dots r}$ such that $T \approx \sum_{k=1}^r w_k u_k^1 \otimes u_k^2 \otimes \cdots \otimes
u_k^n$}
 \While{stopping criterion}{
    \For{$l = 1, \dots, r$}{
        \For{$j = 1, \dots, n$}{
            $u_l^j = {(T - \sum_{k \neq l}^r u_k^1 \otimes u_k^2 \otimes \cdots \otimes u_k^n)(u_{l}^1, \dots, u_{l}^{j-1}, I, u_{l}^{j+1}, \dots, u_{l}^n)}$\;
            $u_l^j = \frac{u_l^j}{\|u_l^j\|}$\;
        }
        $w_l = (T - \sum_{k \neq l}^r u_k^1 \otimes u_k^2 \otimes \cdots \otimes u_k^n)( u_l^1, \dots, u_l^n)$\;
    }
 }
 \caption{Alternating minimization}\label{algo:altmin}
\end{algorithm}

\section{Policy improvement}\label{section:policyimprovement}
Policy improvement is a well known method in reinforcement learning to compute optimal policies with respect to the MDP parameters $T$ and $R$. In model based reinforcement learning algorithms, this is used to compute a good policy after estimating $T$ and $R$ with estimates $\hat T$ and $\hat R$. If the estimates are close enough, the optimal policy with respect to $\hat T$, $\hat R$ will also perform well on the actual MDP described by $T, R$.

Computing the optimal policy uses the $Q$-function, which maps each state $s$ and each action $a$ to the expected reward of executing action $a$ in state $s$ and following policy $\pi$ afterwards. This can recursively be written as 

$$Q^\pi(s, a) = R(s, a) + \gamma \sum_{s' \in S}T(s' | a, s)\sum_{a' \in A}\pi(a' | s')Q^\pi(s' , a')$$

The value function maps each state to the expected future reward in the state when following policy $\pi$. This can be computed from the $Q$-function via 
$$V^\pi(s) = \sum_{a \in A}Q(s, a)\pi(a | s)$$
The main idea of policy improvement is to iteratively select states for which $V^\pi(s) < \max_{a \in A} Q^\pi(s, a)$. This means that there exists an action which achieves better reward than the current policy, so the policy $\pi$ is updated to use the better action instead. After this, since the policy has changed, $Q^\pi$ and $V^\pi$ need to be computed again to repeat this process. This is guaranteed to converge to an optimal policy eventually on finite state and action spaces.

\begin{algorithm}[H]
\SetAlgoLined
\KwInput{Starting policy $\pi_0$}
\KwResult{Improved policy $\pi$}
 Set $i = 0$\;
 Compute $Q^{\pi_{i}}$ and $V^{\pi_{i}}$\;
 \While{there exist $s \in S$ such that $V^{\pi_i}(s) < \max_a Q^\pi(s, a)$}{
    Pick $s$ such that $V^{\pi_i}(s) < \max_a Q^{\pi_i}(s, a)$\;
    Let $A^*_{s, \pi_i} = \argmax_a Q^\pi(s, a)$\;
    Define a new policy: $\pi_{i+1} = \pi_i$\;
    Set $\pi_{i+1}(A^*_{s, \pi_{i}} | s) = 1$\;
    $i = i+1$\;
    Compute $Q^{\pi_{i}}$ and $V^{\pi_{i}}$\;
 }
 \caption{Policy improvement}\label{algo:policyimprovement}
\end{algorithm}
\section{Tensor completion} \label{section:appendixtensorcompletion}
To modify algorithm \ref{algo:altrank1updates} for tensor completion, alternating minimization (algorithm \ref{algo:altmin}) can be modified to solve $$\argmin_{u_l^j \in \mathbb{R}^{d_j}} \|\Omega \cdot T - \Omega \cdot \sum_{k=1}^r w_k u_k^1 \otimes u_k^2 \otimes \cdots \otimes u_k^n\|_F$$
at each iteration, instead of the usual problem 
$$\argmin_{u_l^j \in \mathbb{R}^{d_j}} \|T - \sum_{k=1}^r w_k u_k^1 \otimes u_k^2 \otimes \cdots \otimes u_k^n\|_F.$$
This leads to the update 
$$u_l^j = \frac{( \Omega \cdot T - \Omega \cdot \sum_{k \neq l}^rw_k u_k^1 \otimes u_k^2 \otimes \cdots \otimes u_k^n)(u_{l}^1, \dots, u_{l}^{j-1}, I, u_{l}^{j+1}, \dots, u_{l}^n)}{\Omega( u_{l}^1 \cdot u_{l}^1, \dots, u_{l}^{j-1}\cdot u_{l}^{j-1}, I, u_{l}^{j+1}\cdot u_{l}^{j+1}, \dots, u_{l}^n \cdot u_{l}^n)}$$
instead of the usual update displayed in algorithm \ref{algo:altmin}. Similarly, the update for the weights is given by

$$w_l = \frac{( \Omega \cdot T - \Omega \cdot \sum_{k \neq l}^r w_ku_k^1 \otimes u_k^2 \otimes \cdots \otimes u_k^n)(u_{l}^1, \dots, u_{l}^n)}{\Omega( u_{l}^1 \cdot u_{l}^1, \dots, u_{l}^n \cdot u_{l}^n)}.$$
This modification is inspired by the tensor completion method in \cite{tensorcompletion1}, which is proven to work for symmetric orthogonal tensors.
\section{Tensor generation details}\label{section:tensorgeneration}
The target reward and transition tensors in the experiments are generated by algorithm \ref{algo:generatetensors}. The weights are chosen to be $w = \texttt{linspace}(0.1, 1)$ for the reward tensor in the first experiment.

\begin{algorithm}\caption{Tensor generation}\label{algo:generatetensors}
\KwInput{Desired shape $(d_1, \dots, d_n)$ and rank $r$, weights $w \in \mathbb{R}^r$}
\KwResult{Rank $r$ tensor with dimensions $d_1, \dots, d_n$}
\For{$i = 1, \dots, n$}{
    \For{$l = 1, \dots, r$}{
        $u_{i}^l \sim \mathcal{N}(0, I_{d_i})$\;
        $u_{i}^l = \frac{u_{i}^l}{\|u_{i}^l\|}$\;
    }
}
$T = \sum_{l=1}^r w_l u_{1}^l \otimes \cdots \otimes u_{n}^l$\;
\end{algorithm}

A complication is that the transition tensor must satisfy $\sum_{s'}T_sa_1a_2a_3s' = 1$. It is difficult to directly generate a tensor of fixed rank with this property, and normalizing a tensor by setting $$T'_{s_1a_1a_2a_3s_2} = \frac{T_{s_1a_1a_2a_3s_2}}{\sum_{s'}T_{s_1a_1a_2a_3s'}}$$
changes the rank of the tensor. This is overcome by iteratively normalizing, computing a new decomposition of higher than the desired rank, and then truncating it to the desired rank. Repeating this as shown in algorithm \ref{algo:generatetransitiontensors} appears to converge in practice, enabling the generation of a valid transition tensor that is arbitrarily close to to a tensor of desired rank.

\begin{algorithm}\caption{Transition tensor generation}\label{algo:generatetransitiontensors}
\KwInput{Desired shape $(d_1, \dots, d_n)$ and rank $r$, tolerance $\epsilon$}
\KwResult{Approximately rank $r$ transition tensor with dimensions $d_1, \dots, d_n$}
$T = \texttt{GenerateTensor}( (d_1, \dots, d_n), r, w)$\;
\While{ $\|\texttt{Normalize}(T) - T\|_F > \epsilon$}{
    $T = \texttt{Normalize(T)}$\;
    $T = \texttt{TensorDecomp(T, 2r)}$\;
    $T = \texttt{Truncate}(T, r)$\;
}
$T = \texttt{Normalize}(T)$\;
\end{algorithm}

\section{Extra experiment figures}
\begin{figure}
    \centering
    \includegraphics[width=\textwidth]{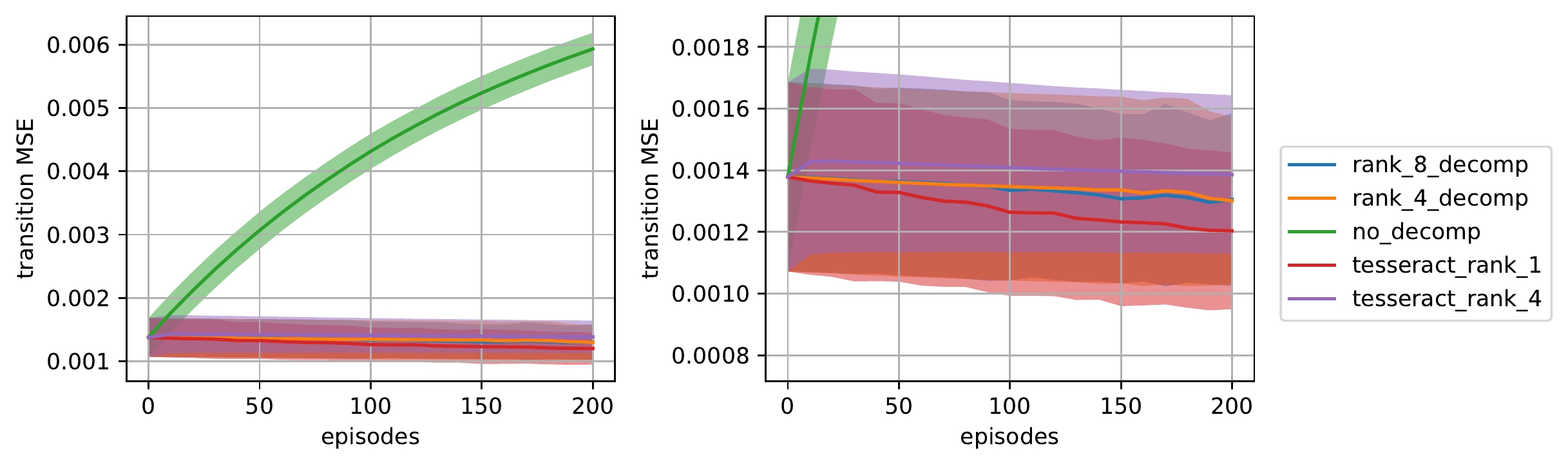}
    \caption{Mean squared error of the estimate of the transition tensor after training a number of episodes in the state degeneracy MDP. The right plot is a zoomed in version of the left plot.}
    \label{fig:degen_transitionerrors}
\end{figure}

\begin{figure}[H]
     \centering
     \begin{subfigure}[b]{0.45\textwidth}
    \centering
    \includegraphics[width=\textwidth]{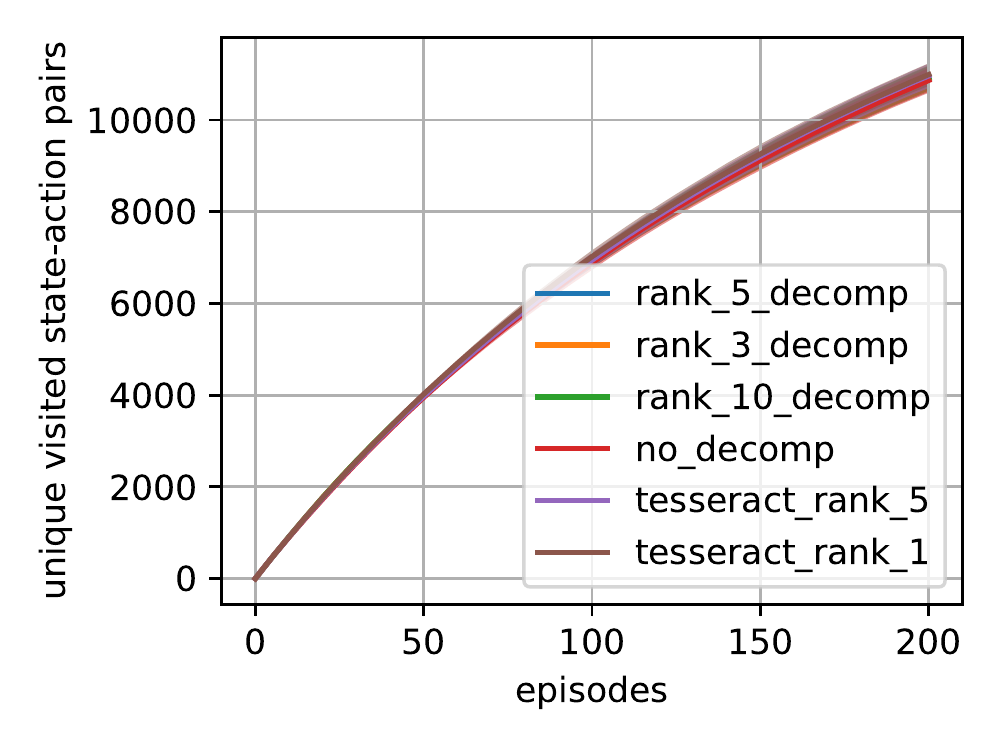}
    \caption{Number of unique states visited after training for a number of episodes in the first experiment. The MDP has 20000 unique states in total}
    \label{fig:predef_visited}
     \end{subfigure}
     \hfill
     \begin{subfigure}[b]{0.45\textwidth}
    \centering
    \includegraphics[width=\textwidth]{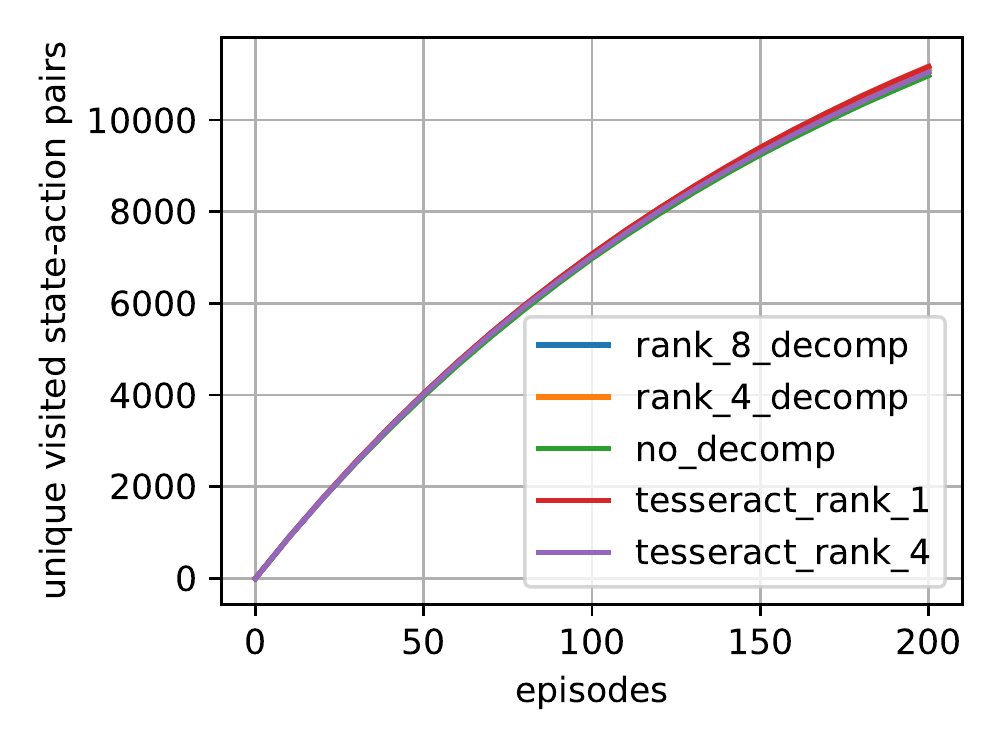}
    \caption{Number of unique states visited after training for a number of episodes in the state degeneracy experiment. The total number of state-action pairs in the MDP is 128000}
    \label{fig:degen_visited}
     \end{subfigure}
\end{figure}
\end{document}